\documentclass[10pt,twocolumn,letterpaper]{article}
\usepackage{blindtext}
\usepackage{cvpr}
\usepackage{times}
\usepackage{epsfig}
\usepackage{graphicx}
\usepackage{amsmath}
\usepackage{enumitem}
\usepackage{amssymb}
\usepackage[pagebackref=true,breaklinks=true,letterpaper=true,colorlinks,bookmarks=false]{hyperref}

\cvprfinalcopy 

\def\cvprPaperID{3139} 

\ifcvprfinal\fi
\begin{document}

\title{Image Generation from Layout}

\author{Bo Zhao \quad \quad Lili Meng \quad \quad Weidong Yin \quad \quad Leonid Sigal\\
University of British Columbia  \quad Vector Institute\\
{\tt\small \{bzhao03, menglili, wdyin, lsigal\}@cs.ubc.ca}
}

\maketitle

\begin{abstract}
Despite significant recent progress on generative models, controlled generation of images depicting multiple and complex object layouts is still a difficult problem. 
Among the core challenges are the diversity of appearance a given object may possess and, as a result, exponential set of images consistent with a specified layout. 
To address these challenges, we propose a novel approach for layout-based image generation; we call it Layout2Im.
Given the coarse spatial layout (bounding boxes + object categories), our model can generate a set of realistic images which have the correct objects in the desired locations.
The representation of each object is disentangled into a specified/certain part (category) and an unspecified/uncertain part (appearance).
The category is encoded using a word embedding and the appearance is distilled into a low-dimensional vector sampled from a normal distribution.  
Individual object representations are composed together using convolutional LSTM, to obtain an encoding of the complete layout, and then decoded to an image.  
Several loss terms are introduced to encourage accurate and diverse image generation. 
The proposed Layout2Im model significantly outperforms the previous state-of-the-art, boosting the best reported inception score by 24.66\% and 28.57\% on the very challenging COCO-Stuff and Visual Genome datasets, respectively.
Extensive experiments also demonstrate our model's ability to generate complex and diverse images with many objects.

\end{abstract}

\section{Introduction}

Image generation of complex realistic scenes with multiple objects and desired layouts is one of the core frontiers for computer vision. Existence of such algorithms would not only inform our designs for inference mechanisms, needed for visual understanding, but also provide practical application benefits in terms of automatic image generation for artists and users. In fact, such algorithms, if successful, may replace visual search and retrieval engines in their entirety. Why search the web for an image, if you can create one to user specification? 

\begin{figure}[!t]
    \begin{center}
        \includegraphics[width=1\linewidth]{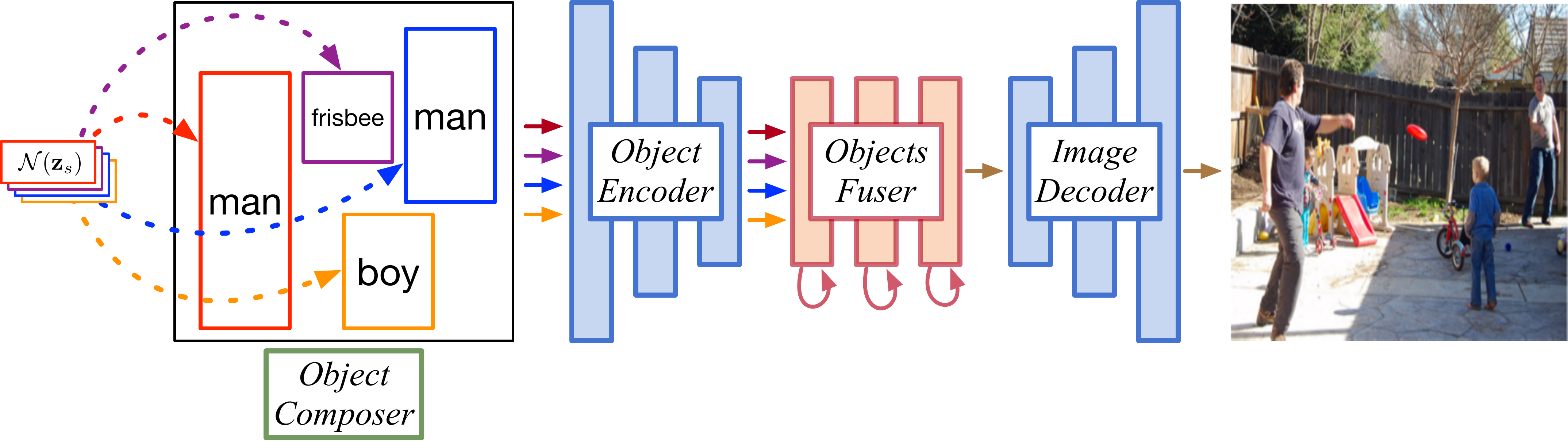}
    \end{center}
    \caption{{\bf Image generation from layout.} Given the coarse layout (bounding boxes + object categories), the proposed Layout2Im model samples the appearance of each object from a normal distribution, and transforms these inputs into a real image by a serial of components. Please refer to Section~\ref{sec:method} for a detailed explanation.}
    \label{fig:task_example}
\end{figure}

For these reasons, image generation algorithms have been a major focus of recent research. Of specific relevance are approaches for text-to-image~\cite{hong2018inferring, karacan2016learning, mansimov2015generating, reed2017parallel,tan2018text2scene,zhang2017stackgan} generation. 
By allowing users to describe visual concepts in natural language, text-to-image generation provides natural and flexible interface for conditioned image generation.
However, existing text-to-image approaches exhibit two drawbacks: (i) most approaches can only generate plausible results on simple datasets such as cats~\cite{Zhang2008}, birds~\cite{WelinderEtal2010} or flowers~\cite{flowers}. Generating complex, real-world images such as those in COCO-Stuff~\cite{caesar2016coco} and Visual Genome~\cite{krishna2017visual} datasets remains a challenge; (ii) the ambiguity of textual description makes it more difficult to constrain complex generation process, \eg, locations and sizes of different objects are usually not given in the description. 

Scene graphs are powerful structured representations that encode objects, their attributes and relationships. In~\cite{Johnson2018} an approach for generating complex images with many objects and relationships is proposed by conditioning the generation on scene graphs. It addresses some of the aforementioned challenges. However, scene graphs are difficult to construct for a layman user and lack specification of core spatial properties, \eg, object size / position.



To overcome these limitations, we propose to generate complicated real-world images from layouts, as illustrated in Figure~\ref{fig:task_example}. By simply specifying the coarse layout (bounding boxes + categories) of the expected image, our proposed model can generate an image which contains the desired objects in the correct locations. It is much more controllable and flexible to generate an image from layout than textual description.

With the new task comes new challenges. 
First, image generation from layout is a difficult one-to-many problem. 
Many images could be consistent with a specified layout; same layout may be realized by different appearance of objects, or even their interactions (\eg, a person next to the frisbee may be throwing it or be a bystander, see Figure~\ref{fig:task_example}).
Second, the information conveyed by a bounding box and corresponding label is very limited. The actual appearance of the object displayed in an image is not only determined by its category and location, but also its interactions and consistency with other objects.
Moreover, spatially close objects may have overlapping bounding boxes. 
This leads to additional challenges of ``separating'' which object should contribute to individual pixels. 
A good generative model should take all these factors and challenges into account implicitly or explicitly. 

We address these challenges using a novel variational inference approach.  
The representation of each object in the image is explicitly disentangled into a specified/certain part (category) and an unspecified/uncertain part (appearance).
The category is encoded using a word embedding and the appearance is distilled into a low-dimensional vector sampled from a normal distribution. 
Based on this representation and specification of object bounding box, we construct a feature map for each object. These feature maps are then composed using convolutional LSTM into a hidden feature map for the entire image, which subsequently is decoded into an output image. 
This set of modelling choices makes it easy to generate different and diverse images by sampling the appearance of individual objects, and/or adding, moving or deleting objects from the layout.
Our proposed model is end-to-end learned using a loss that consists of a number of objectives.
Specifically, a pair of discriminators are designed to discriminate the overall generated image and the generated objects within their specified bounding boxes, as real or fake. In addition, object discriminator is also trained to classify the categories of generated objects. 

\noindent
{\bf Contributions.}
Our contributions are three-fold:
(1) We propose a novel approach for generating images from coarse layout (bounding boxes + object categories). This provides a flexible control mechanism for image generation. 
(2) By disentangling the representation of objects into a category and (sampled) appearance, our model is capable of generating a diverse set of consistent images from the same layout.
(3) We show qualitative and quantitative results on COCO-Stuff~\cite{caesar2016coco} and Visual Genome~\cite{krishna2017visual} datasets, demonstrating our model's ability to generate complex images with respect to object categories and their layout (without access to segmentation masks~\cite{hong2018inferring,Johnson2018}). 
We also preform comprehensive ablations to validate each component in our approach.


\section{Related Work}
\label{sec:related_work}
\paragraph{Conditional Image Generation.} Conditional image generation approaches generate images conditioned on additional input information, including entire source image~\cite{isola2017image,liu2017unsupervised, Pathak2016, Yang2017, Zhao2018, zhu2017unpaired, zhu2017toward}, sketches~\cite{isola2017image, Sangkloy2017, wang2017high, Xian2018,zhu2017toward}, scene graphs~\cite{Johnson2018}, dialogues~\cite{kim2017codraw, sharma2018chatpainter} and text descriptions~\cite{mansimov2015generating, reed2017parallel,  tan2018text2scene, zhang2017stackgan}. Variational Autoencoders (VAEs)~\cite{kingma2013auto, mansimov2015generating, sohn2015learning}, autoregressive models~\cite{oord2016pixel, van2016conditional} and GANs~\cite{isola2017image, mirza2014conditional, wang2017high, zhu2017unpaired} are powerful tools for conditional image generation and have shown promising results.  
However, many previous generative models~\cite{isola2017image,Pathak2016,Sangkloy2017,Xian2018,Yang2017,zhu2017unpaired} tend to largely ignore the random noise vector when conditioning on the same relevant context, making the generated images very similar to each other. 
By enforcing the bijection mapping between the latent and target space, BicycleGAN~\cite{zhu2017toward} pursues the diversity of generated images from a same input. Inspired by this idea, in our paper, we also explicitly regress the latent codes which are used to generate the different objects.


\vspace{-0.15in}
\paragraph{Image Generation from Layout.} 
The use of layout in image generation is a relative novel task. 
In prior works, it is usually served as an intermediate representation between other input sources (\eg, text~\cite{hong2018inferring} or scene graphs~\cite{Johnson2018}) and the output images, or as a complementary feature for image generation based on context (\eg, text~\cite{karacan2016learning, reed2016learning, tan2018text2scene}, shape and lighting~\cite{dosovitskiy2015learning}). 
In~\cite{hong2018inferring,Johnson2018}, instead of learning a direct mapping from textual description/scene graph to an image, the generation process is decomposed into multiple individual steps. They first construct a semantic layout (bounding boxes + object shapes) from the input, and then convert it to an image using an image generator. 
Both of them can generate an image from a coarse layout together with textual description/scene graph. However~\cite{hong2018inferring} requires detailed object instance segmentation masks to train its object shape generator. Getting such segmentation masks for large scale datasets is both time-consuming and and labor-intensive.
Different from~\cite{hong2018inferring} and~\cite{Johnson2018}, we use the coarse layout without instance segmentation mask as a fundamental input modality for diverse image generation.

\begin{figure*}[!ht]
  \centering
  \includegraphics[width=\linewidth]{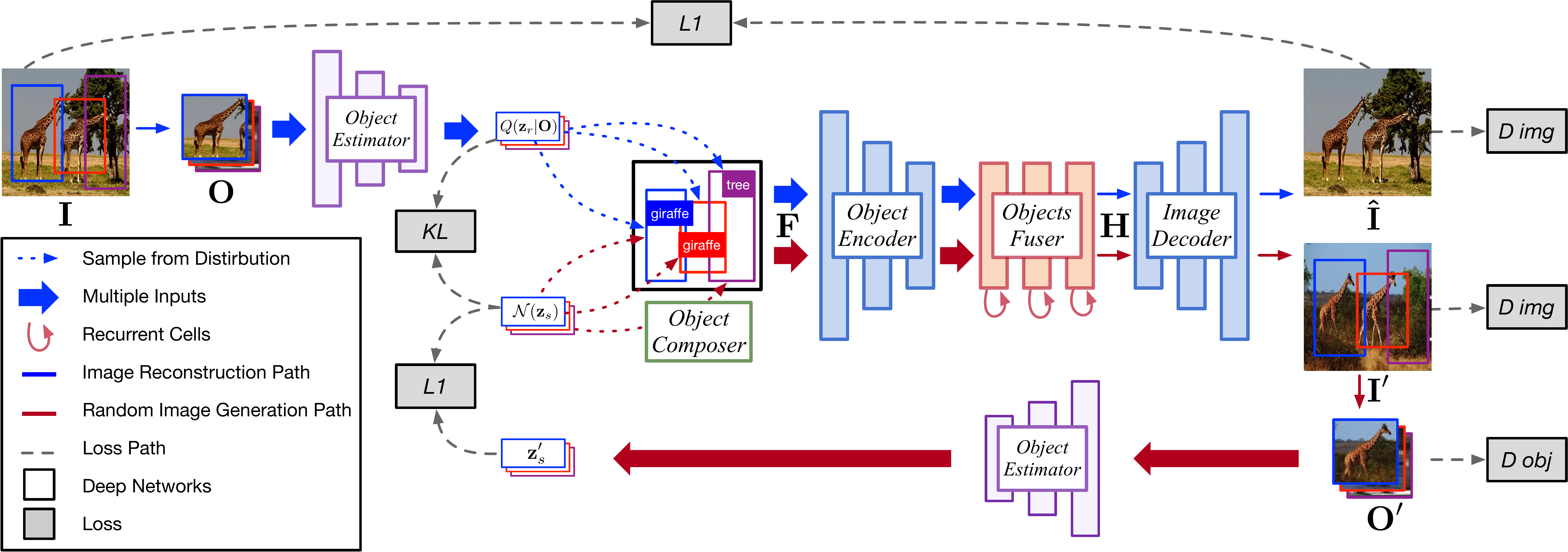}
  \caption{{\bf Overview of our Layout2Im network} for generating images from layout during training. The inputs to the model are the ground truth image with its layout. The objects are first cropped from the input image according to their bounding boxes, and then processed with the object estimator to predict a latent code for each object. After that, multiple object feature maps are prepared by the object composer based on the latent codes and layout, and processed with the object encoder, objects fuser and image decoder to reconstruct the input image. Additional set of latent codes are also sampled from a normal distribution to generate a new image. Finally, objects in generated images are used to regress the sampled latent codes. The model is trained adversarially against a pair of discriminators and a number of objectives.}
  \vspace{-0.2in}
  \label{fig:pipeline}
\end{figure*}

\vspace{-0.15in}
\paragraph{Disentangled Representations.}
Many papers~\cite{chen2016infogan,Cheung2015,Denton2017,Lai2017,lee2018diverse,Ma2018,Mathieu2016,Murez2018} have tried to learn disentangled representations as part of image generation. Disentangled representations model different factors of data variations, such as class-related and class-independent parts~\cite{Cheung2015,Lai2017,lee2018diverse,Mathieu2016,Murez2018}. By manipulating the disentangled representations, images with different appearances can be generated easily.
In~\cite{Ma2018}, three factors (foreground, background and pose) are disentangled explicitly when generating person image. 
InfoGAN~\cite{chen2016infogan}, DrNet~\cite{Denton2017} and DRIT~\cite{lee2018diverse} learn the disentangled representations in an unsupervised manner, either by maximizing the mutual information~\cite{chen2016infogan} 
or adversarial losses~\cite{Denton2017, lee2018diverse}. 
In our work, we explicitly separate the representation of each object into a category-related and an appearance-related parts, and only the bounding boxes and category labels are used during both training and testing.


\section{Image Generation from Layout}
\label{sec:method}


The overall {\bf training} pipeline of the proposed approach is illustrated in Figure~\ref{fig:pipeline}.
Given a ground-truth image $\mathbf{I}$ and its corresponding layout $\mathbf{L}$, where $\mathbf{L}_i = (x_i, y_i, h_i, w_i)$ containing the top-left  coordinate, height and width of the bounding box, our model first samples two latent codes $\mathbf{z}_{ri}$ and $\mathbf{z}_{si}$ for each object instance $\mathbf{O}_i$. 
The $\mathbf{z}_{ri}$ is sampled from the posterior $Q(\mathbf{z}_{r}|\mathbf{O}_i)$ conditioned on
object $\mathbf{O}_i$ cropped from the input image according to $\mathbf{L}_i$. The $\mathbf{z}_{si}$ is sampled from a normal prior distribution $\mathcal{N}(\mathbf{z}_{s})$.
Each object $\mathbf{O}_i$ also has a word embedding $\mathbf{w}_i$, which is an embedding of its category label $y_i$.
Based on the latent codes $\mathbf{z}_i \in \{ \mathbf{z}_{ri}, \mathbf{z}_{si} \}$, word embedding $\mathbf{w}_i$, and layout $\mathbf{L}_i$, multiple object feature maps $\mathbf{F}_i$ are constructed, and then fed into the object encoder and the objects fuser sequentially, generating a fused hidden feature map $\mathbf{H}$ containing information from all specified objects.
Finally, an image decoder $D$ is used to reconstruct, $\hat{\mathbf{I}} = D(\mathbf{H})$, the input ground-truth image $\mathbf{I}$ and generate a new image $\mathbf{I}'$, simultaneously; the former comes from $\mathbf{z}_{r} = \{ \mathbf{z}_{ri} \}$ and the latter from $\mathbf{z}_{s} = \{ \mathbf{z}_{si} \}$. Notably, both resulting images match the training image input layout.
To make the mapping between the generated object $\mathbf{O}'_i$ and the sampled latent code $\mathbf{z}_{si}$ consistent, we make the object estimator regress the sampled latent codes $\mathbf{z}_{si}$ based on the generated object $\mathbf{O}'_i$ in $\mathbf{I}'$ at locations $\mathbf{L}_i$.
To train the model adversarially, we also introduce a pair of discriminators, $D_\mathrm{img}$ and $D_\mathrm{obj}$, to classify the results at image and object level as being real or fake.
%


Once the model is trained, it can generate a new image from a layout by sampling object latent codes from the normal prior distribution $\mathcal{N}(\mathbf{z}_{s})$ as illustrated in Figure~\ref{fig:task_example}.

\subsection{Object Latent Code Estimation}
Object latent code posterior distributions are first estimated from the ground-truth image, and used to sample object latent code $\mathbf{z}_{ri} \sim Q(\mathbf{z}_{ri} | \mathbf{O}_i) =  \mathcal{N}(\mu(\mathbf{O}_i), \sigma(\mathbf{O}_i))$. 
These object latent codes model the ambiguity in object appearance in the ground-truth image, and play important roles in reconstructing the input image later.

\begin{figure}[!t]
    \begin{center}
        \includegraphics[width=1\linewidth]{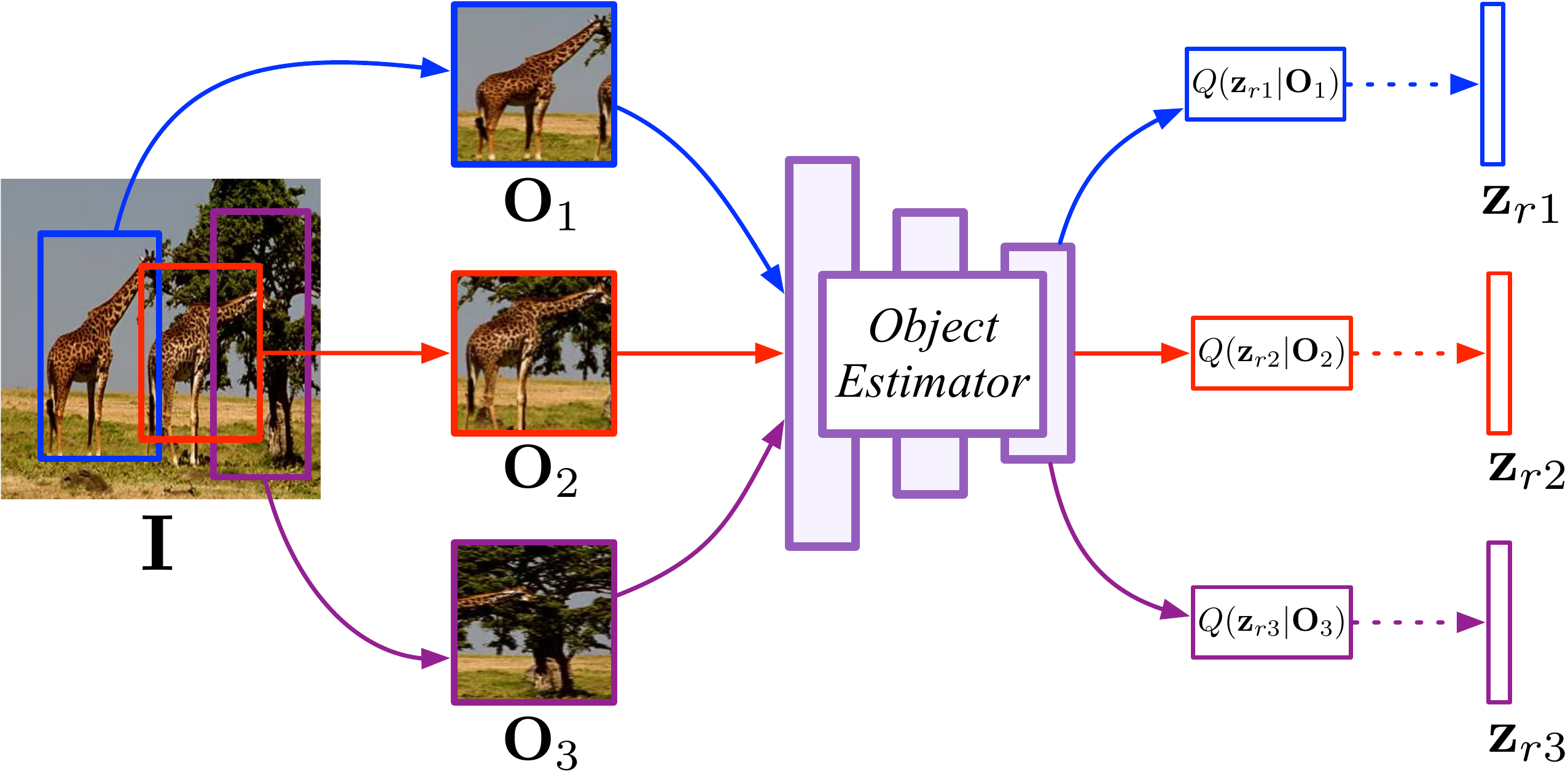}
    \end{center}
    \caption{{\bf Object latent code estimation.} Given the input image and its layout, the objects are first cropped and resized from the input image. Then the object estimator predicts a distribution for each object from the object crops, and multiple latent codes are sampled from the estimated distribution.}
    \label{fig:latent_code_estimation}
    \vspace{-0.1in}
\end{figure}

Figure~\ref{fig:latent_code_estimation} illustrates the object latent code estimation process.
First, each object $\mathbf{O}_i$ is cropped, from the input image $\mathbf{I}$ according to its bounding box $\mathbf{L}_i$, and then resized to fit the input dimensionality of object estimator using bilinear interpolation. 
The resized object crops are fed into an object estimator which consists of several convolutional layers and two fully-connected layers. The object estimator predicts the mean and variance of the posterior distribution for each input object $\mathbf{O}_i$. 
Finally, the predicted mean and variance are used to sample a latent code $\mathbf{z}_{ri}$ for the input object $\mathbf{O}_i$.
We sample latent code for every object in the input image.

\subsection{Object Feature Map Composition}

Given the object latent code $\mathbf{z}_i \in \mathbb{R}^{m}$ sampled from either posterior or the prior ( $\mathbf{z}_i \in \{ \mathbf{z}_{ri}, \mathbf{z}_{si} \}$), object category label $y_i$ and corresponding bounding box information $\mathbf{L}_i$, the object composer module constructs a feature map $\mathbf{F}_i$ for each object $\mathbf{O}_i$. Each feature map $\mathbf{F}_i$ contains a region corresponding to $\mathbf{L}_i$ filled with the disentangled representation of that object, consisting of object identity and appearance.

\begin{figure}[!t]
    \begin{center}
        \includegraphics[width=1\linewidth]{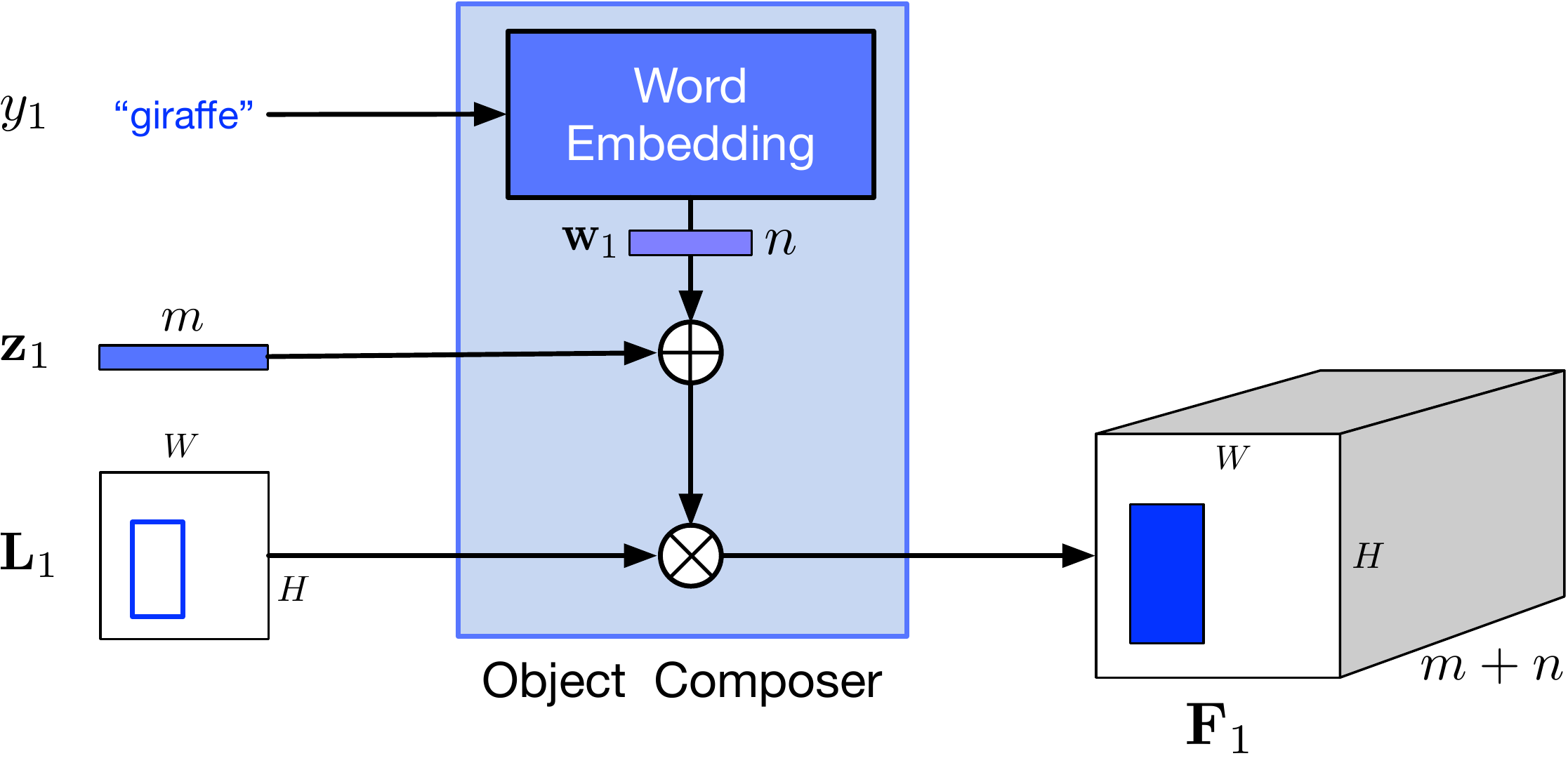}
    \end{center}
    \caption{{\bf Object feature map composition.} The object category is first encoded by a word embedding. Then the object feature map is simply composed by filling the region within the object bounding box with the concatenation of category embedding and latent code. The rest of the feature map are all zeros. Symbol $\bigoplus$ stands for the vector concatenation, and $\bigotimes$ means replicating object representation within a bounding box.}
    \label{fig:layout_composition}
    \vspace{-0.1in}
\end{figure}

Figure~\ref{fig:layout_composition} illustrates this module.
The object category label $y_i$ is first transformed to a corresponding word vector embedding $\mathbf{w}_i \in \mathbb{R}^{n}$, and then concatenated with the object latent vector $\mathbf{z}_i$. 
This results in the representation of the object which has two parts: object embedding and object latent code. Intuitively, the object embedding encodes the identity of the object, while the latent code encodes the appearance of a specific instance of that object. Jointly these two components encode sufficient information to reconstruct a specific instance of the object in an image. 
The object feature map $\mathbf{F}_i$ is composed by simply filling the region within its bounding box with this object representation $(\mathbf{w}_i, \mathbf{z}_i) \in \mathbb{R}^{m+n}$.
For each  tuple $<y_i, \mathbf{z}_i, \mathbf{L}_i>$ encoding object label, latent code and bounding box, we compose an object feature map $\mathbf{F}_i$. 
These object feature maps are downsampled by an object encoder network which contains several convolutional layers. 
Then an object fuser module is used to fuse all the downsampled object feature maps, generating a hidden feature map $\mathbf{H}$.


\subsection{Object Feature Maps Fusion}
Since the result image will be decoded from it, a good hidden feature map $\mathbf{H}$ is crucial to generating a realistic image. 
The properties of a good hidden feature map can be summarized as follows:
(i) it should encode all object instances in the desired locations; 
(ii) it should coordinate object representations based on other objects in the image;
(iii) it should be able to fill the unspecified regions, \eg, background, by implicitly reasoning about plausibility of the scene with respect to the specified objects. 

To satisfy these requirements, we choose a multi-layer convolutional Long-Short-Term Memory (cLSTM) network~\cite{Shi2015} to fuse the downsampled object feature maps $\mathbf{F}$. 
Different from the traditional LSTM~\cite{Hochreiter1997}, the hidden states and cell states in cLSTM are both feature maps rather than vectors. The computation of different gates are also done by convolutional layers. Therefore, cLSTM can better preserve the spatial information compared with the traditional vector-based LSTM.  
The cLSTM acts like an encoder to integrate object feature maps one-by-one, and the last output of the cLSTM is used as the fused hidden layout $\mathbf{H}$, which incorporates the location and category information of all objects. Please refer to the supplementary material for the structure of multi-layer cLSTM networks.

\subsection{Image Decoder}
Given the fused image hidden feature map $\mathbf{H}$, image decoder is tasked with generating a result image.
As shown in Figure~\ref{fig:pipeline}, there are two paths (blue and red) in the networks. They differ in latent code estimation. 
The blue path reconstructs the input image using the object latent codes $\mathbf{z}_r$ sampled from the posteriors $Q(\mathbf{z}_r | \mathbf{O})$ that are conditioned on the objects $\mathbf{O}$ in the input image $\mathbf{I}$, while in the red one, the latent codes $\mathbf{z}_s$ are directly sampled from prior distributions $\mathcal{N}(\mathbf{z}_s)$. 
As a result, two images are generated, \ie, $\mathbf{\hat{I}}$ and $\mathbf{I}'$, through the red and blue paths, respectively. Althrough they may differ in appearance, both of them share the same layout.



\subsection{Object Latent Code Regression}
To explicitly encourage the consistent connection between the latent codes and outputs, our model also tries to recover the random sampled latent codes from the objects generated along the red path. One can think of this as an inference network for the latent codes. This helps prevent a many-to-one mapping from the latent code to the output during training, and as a result, produces more diverse results.

To achieve this, we use the same input object bounding boxes $\mathbf{L}$ to crop the objects $\mathbf{O}'$ in the {\em generated} image $\mathbf{I}'$. 
The resized $\mathbf{O}'$ are then sent to an object latent code estimator (which shares weights with the one used in image reconstruction path), getting the estimated mean and variance vectors for the generated objects. 
We directly use the computed mean vectors, as the regressed latent codes $\mathbf{z}'_s$, and compare them with the sampled ones $\mathbf{z}_s$, for all objects. 

\subsection{Image and Object Discriminators}
To make the generated images realistic, and the objects recognizable, we adopt a pair of discriminators $D_{\mathrm{img}}$ and $D_{\mathrm{obj}}$.
The discriminator is trained to classify an input $x$ or $y$ as real or fake by maximizing the objective~\cite{Goodfellow2014}:
\begin{align}
	\label{eq:adv_loss}
	\mathcal{L}_\mathrm{GAN} = \underset{x\sim p_\mathrm{real}}{\mathbb{E}}\log D(x)+ \underset{y\sim p_\mathrm{fake}}{\mathbb{E}}\log(1-D(y)),
\end{align}
where $x$ represents the real images and $y$ represents the generated ones.
Meanwhile, the generator networks are trained to minimizing $\mathcal{L}_\mathrm{GAN}$.
The image discriminator $D_{\mathrm{img}}$ is applied to input images $\mathbf{I}$, reconstructed images $\hat{\mathbf{I}}$ and sampled images $\mathbf{I}'$, classifying them as real or fake.

The object discriminator $D_{\mathrm{obj}}$ is designed to assess the quality and category of the real objects $\mathbf{O}$, reconstructed objects $\mathbf{\hat{O}}$ and sampled objects $\mathbf{O}'$ at the same time.
In addition, since $\mathbf{\hat{O}}$ and $\mathbf{O}'$ are cropped from the reconstructed/sampled images according to the input bounding boxes $\mathbf{L}$, $D_{obj}$ also encourages the generated objects to appear in their desired locations. 

 

\subsection{Loss Function}
We end-to-end train the generator network and two discriminator networks in an adversarial manner. The generator network, with all described components, is trained to minimize the weighted sum of six losses:

\begin{itemize}[leftmargin=*]
\setlength{\itemsep}{0pt}
  \item \textbf{KL Loss} $\mathcal{L}_{\mathrm{KL}} = \sum_{i=1}^{o}\mathbb{E}[\mathcal{D}_{\mathrm{KL}}(Q(\mathbf{z}_{ri}|\mathbf{O}_i)||\mathcal{N}(\mathbf{z}_r))]$ computes the KL-Divergence between the distribution $Q(\mathbf{z}_r|\mathbf{O})$ and the normal distribution $\mathcal{N}(\mathbf{z}_r)$, where $o$ is the number of objects in the image/layout. 
  \item \textbf{Image Reconstruction Loss} $\mathcal{L}_1^{\mathrm{img}} = ||\mathbf{I}-\hat{\mathbf{I}}||_1$ penalizes the $\mathcal{L}_1$ difference between ground-truth image $\mathbf{I}$ and reconstructed image $\mathbf{\hat{I}}$. 
  \item \textbf{Object Latent Code Reconstruction Loss} $\mathcal{L}_1^{\mathrm{latent}} = \sum_{i=1}^{o}||\mathbf{z}_{si}-\mathbf{z'}_{si}||_1$ penalizes the $\mathcal{L}_1$ difference between the randomly sampled $\mathbf{z}_{s} \sim N(\mathbf{z}_s)$ and the re-estimated $\mathbf{z}'_s$ from the generated objects $\mathbf{O}'$.
  \item \textbf{Image Adversarial Loss} $\mathcal{L}_{\mathrm{GAN}}^{\mathrm{img}}$ is defined as in Eq.~\eqref{eq:adv_loss}, where $x$ is the ground truth image $\mathbf{I}$, $y$ is the reconstructed image $\mathbf{\hat{I}}$ and sampled image $\mathbf{I'}$.
  \item \textbf{Object Adversarial Loss} $\mathcal{L}_{\mathrm{GAN}}^{\mathrm{obj}}$ is also defined as in Eq.~\eqref{eq:adv_loss}, where $x$ is the objects $\mathbf{O}$ cropped from the ground truth image $\mathbf{I}$, $y$ are $\mathbf{\hat{O}}$ and $\mathbf{O}'$ cropped from the reconstructed image $\mathbf{\hat{I}}$ and sampled image $\mathbf{I}'$.
  \item \textbf{Auxiliar Classification Loss} $\mathcal{L}_\mathrm{AC}^{\mathrm{obj}}$ from $D_\text{obj}$ encourages the generated objects $\mathbf{\hat{O}}_i$ and $\mathbf{O}'_i$ to be recognizable as their corresponding categories.
\end{itemize}

\noindent
Therefore, the final loss function of our model is defined as:
\begin{align}
    \mathcal{L}  = & \lambda_1 \mathcal{L}_{\mathrm{KL}} + \lambda_2 \mathcal{L}_{1}^{\text{img}} + \lambda_3\mathcal{L}_{1}^{\mathrm{latent}} + \nonumber\\
    &\lambda_4 \mathcal{L}_{\mathrm{adv}}^\mathrm{img} + \lambda_5 \mathcal{L}_{\mathrm{adv}}^\mathrm{obj} + \lambda_6 \mathcal{L}_{\mathrm{AC}}^{\mathrm{obj}}, \nonumber
\end{align}
where, $\lambda_i$ are the parameters balancing different losses.

\subsection{Implementation Details}
We use SN-GAN~\cite{Miyato2018} for stable training. Batch normalization~\cite{ioffe2015batch} and ReLU are used in the object encoder, image decoder, and only ReLU is used in the discriminators (no batch normalization). Conditional batch normalization~\cite{Vries2017} is used in the object estimator to better normalize the object feature map according to its category.
After object fuser, we use six residual blocks~\cite{He2016} to further refine the hidden image feature maps.
We set both $m$ and $n$ to 64. The image and crop size are set to 64 $\times$ 64 and 32 $\times$ 32, respectively. 
The $\lambda_1 \sim \lambda_6$ are set to 0.01, 1, 10, 1, 1 and 1 respectively.

We train all models using Adam~\cite{kingma2014adam} with learning rate of 0.0001 and batch size of 8 for 300,000 iterations; training takes about 3 days on a single Titan Xp GPU.
Full details about our architecture can be found in the supplementary material, and code will be made publicly available.


\section{Experiments}
\label{sec:experiments}
Extensive experiments are conducted to evaluate the proposed Layout2Im network. We first compare our proposed method with previous state-of-the-art models for scene image synthesis, and show its superiority in aspects of realism, recognition and diversity.
Finally, the contributions of each loss for training our model are studied through ablation.

\begin{figure*}[!t]
    \begin{center}
        \includegraphics[width=1\linewidth]{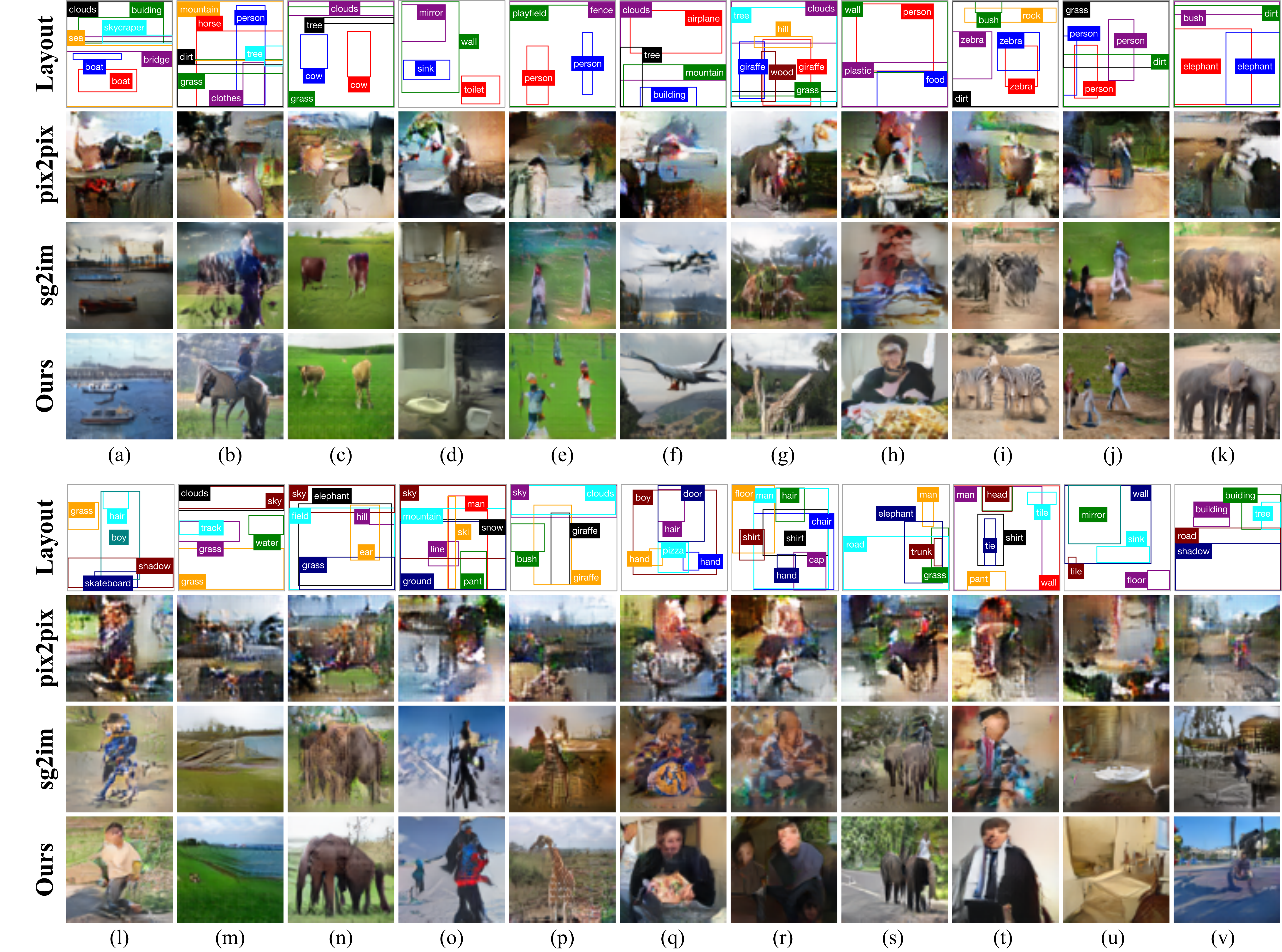}
    \end{center}
    \vspace{-0.1in}
    \caption{\textbf{Examples of 64 $\times$ 64 generated images from complex layouts} on COCO-Stuff (top) and Visual Genome Datasets (bottom) by our proposed method and baselines. For each example, we show the input layout, images generated by pix2pix, sg2im and our method. Please zoom in to see the category of each object. The ground truth images and more examples can be found in the supplementary material.}
    \label{fig:compare_results}
\end{figure*}

\begin{figure*}[!t]
    \begin{center}
        \includegraphics[width=1\linewidth]{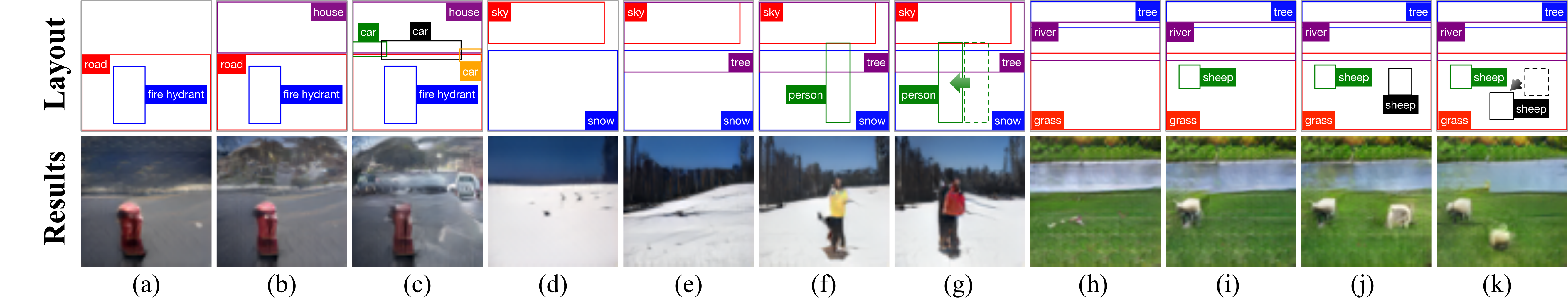}
    \end{center}
    \vspace{-0.1in}
    \caption{\textbf{Example of generated images by adding or moving bounding boxes based on previous layout.} Three groups of images, (a)-(c), (d)-(g) and (h)-(k), are shown. In (g) and (k), original bounding boxes are drawn in dash. Please zoom in to see the category of each object.}
    \vspace{-0.2in}
    \label{fig:adding_results}
\end{figure*}

\begin{figure*}[!t]
    \begin{center}
        \includegraphics[width=1\linewidth]{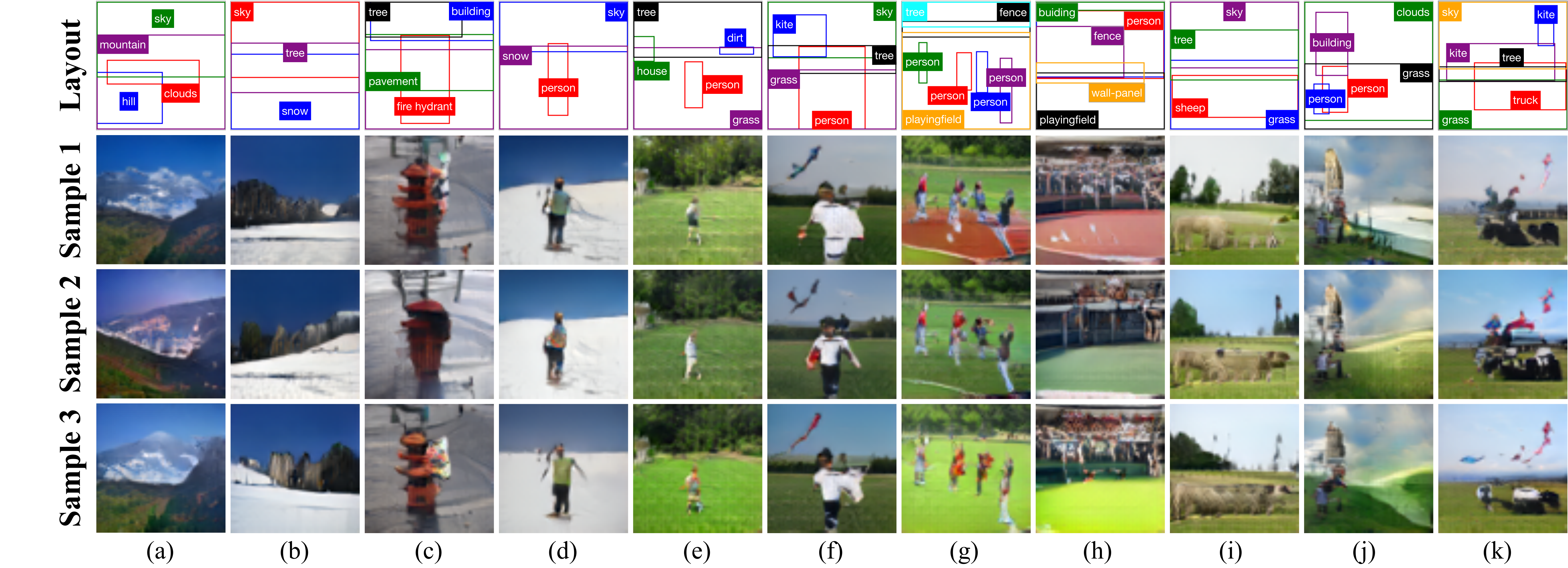}
    \end{center}
    \caption{\textbf{Examples of diverse images generated from same layouts.} For each layout, we sample 3 images. The generated images have different appearances, but sharing the same layout. Please zoom in to see the category of each object.}
    \vspace{-0.2in}
    \label{fig:diverse_results}
\end{figure*}

\subsection{Datasets}

\begin{table}[!t]
\footnotesize
\centering
\begin{tabular}{l|c|c|c|c|l}

\textbf{Dataset}       & \textbf{Train}  & \textbf{Val.}   & \textbf{Test}  & \textbf{\# Obj.} & \textbf{\# Obj. in Image}\\ \hline
COCO~\cite{caesar2016coco}          & 24,972 & 1,024 & 2,048 & 171      & \quad 3 $\sim$ 8               \\ \hline
VG~\cite{krishna2017visual} & 62,565 & 5,506 & 5,088 & 178      & \quad 3 $\sim$ 30             \\ 
\end{tabular}
\vspace{0.05in}
\caption{\textbf{Statistics of COCO-Stuff and Visual Genome dataset.}}
\vspace{-0.2in}
\label{tb:datasets}
\end{table}

The same as previous scene image generation method~\cite{Johnson2018}, we evaluate our proposed model on the COCO-Stuff~\cite{caesar2016coco} and Visual Genome~\cite{krishna2017visual} datasets. We preprocess and split the two datasets the same as that in~\cite{Johnson2018}. Table~\ref{tb:datasets} lists the datasets statistics. Each image in these datasets has multiple bounding boxes annotations with labels for the objects.


\subsection{Baselines}
We compare our approach with two state-of-the-art methods: pix2pix~\cite{isola2017image} and sg2im~\cite{Johnson2018}.

\vspace{0.05in}
\noindent\textbf{pix2pix}~\cite{isola2017image} translates images between two domains. 
In this paper, we define the input domain as feature maps constructed from layout $\mathbf{L}$, and set the real images as the output domain.
We construct the input feature map with the size of $C\times H\times W$ for each layout $\mathbf{L}$, where $C$ is the number of object categories, $H\times W$ is the image size. 
A bounding box $\mathbf{O}_i$ with label $y_i$ will set the corresponding region within $c$-th channel (the channel for category $y_i$) of the feature map to 1 and others are all 0.
The pix2pix model is learned to translate the generated feature maps to real images.

\noindent\textbf{sg2im}~\cite{Johnson2018} is originally trained to generate images from scene graphs. 
However, it can also generate images from layout, simply replacing the predicted layout with ground truth layout. 
We list the Inception Score of sg2im using ground truth layouts as reported in their paper, and generate the results for other comparisons using their released model trained with ground truth layout. In other words, the input and training data for our and sg2im models is identical. 

\subsection{Evaluation Metrics}
Plausible images generated from layout should meet three requirements: be realistic, recognizable and diverse. Therefore we choose four different metrics, Inception Score~(IS)~\cite{salimans2016improved}, Fr\'echet Inception Distance~(FID)~\cite{Heusel2017}, Object Classification Accuracy~(Accu.) and Diversity Score~(DS)~\cite{zhang2018unreasonable}.

\vspace{0.05in}
\noindent\textbf{Inception Score}~\cite{salimans2016improved} is adopted to measure the quality, as well as diversity, of generated images. 
In our paper, we use the pre-trained VGG-net~\cite{simonyan2014very} as the base model to compute the inception scores for our model and the baselines.

\vspace{0.05in}
\noindent\textbf{Fr\'echet Inception Distance}~\cite{Heusel2017} uses 2nd order information of the final layer of the inception model, and calculates the similarity of generated images to real ones. Fr\'echet Inception Distance is more robust to noise than Inception Score.

\vspace{0.05in}
\noindent\textbf{Classification Accuracy} measures the ability to generate recognizable objects, which is an important criteria for our task.
We first train a ResNet-101 model~\cite{He2016} to classify objects. This is done using the real objects cropped and resized from ground truth images in the training set of each dataset. We then compute and report the object classification accuracy for objects in the generated images.

\vspace{0.05in}
\noindent\textbf{Diversity Score} computes the perceptual similarity between two images in deep feature space. 
Different from the inception score which reflects the diversity across the entire generated images, diversity score measures the difference of a pair of images generated from the same input. 
We use the LPIPS metric~\cite{zhang2018unreasonable} for diversity score, and use AlexNet~\cite{krizhevsky2012imagenet} for feature extraction as suggested in the paper.

\begin{table*}[!th]
\centering
\begin{tabular}{l|cc|cc|cc|cc}
                                     & \multicolumn{2}{c|}{\textbf{IS}} & \multicolumn{2}{c|}{\textbf{FID}}         & \multicolumn{2}{c|}{\textbf{Accu.}} & \multicolumn{2}{c}{\textbf{DS}}             \\
\multicolumn{1}{c|}{\textbf{Method}} & \textbf{COCO}          & \textbf{VG}            & \textbf{COCO}      & \textbf{VG}       & \textbf{COCO}            & \textbf{VG}  & \textbf{COCO}            & \textbf{VG}            \\ \hline
Real Images (64 $\times$ 64)         & 16.3 $\pm$ 0.4         & 13.9 $\pm$ 0.5   & - & -     & 55.16              & 49.13             & -                        & -                        \\ \hline
pix2pix~\cite{isola2017image}                              & 3.5 $\pm$ 0.1          & 2.7 $\pm$ 0.02  & 121.97 & 142.86  & 12.06              & 9.20              & 0          & 0 \\
sg2im (GT Layout)~\cite{Johnson2018}                    & 7.3 $\pm$ 0.1          & 6.3 $\pm$ 0.2   & 67.96 & 74.61   & 30.04              & 40.29             & 0.02 $\pm$ 0.01          & 0.15 $\pm$ 0.12          \\ \hline
Ours                                 & \textbf{9.1 $\pm$ 0.1} & \textbf{8.1 $\pm$ 0.1} & \textbf{38.14} & \textbf{31.25} & \textbf{50.84}     & \textbf{48.09}    & \textbf{0.15 $\pm$ 0.06} & \textbf{0.17 $\pm$ 0.09}         
\end{tabular}
\vspace{0.1in}
\caption{\textbf{Performance on COCO and VG in Inception Score~(IS), Fr\'echet Inception Distance~(FID), Object Classification Accuracy~(Accu.) and Diversity Score~(DS).} The output size of all methods is 64 $\times$ 64. We train the pix2pix from scratch, and generate image from the released sg2im model using ground truth layout.}
\vspace{-0.2in}
\label{tb:quantitative_results}
\end{table*}

\subsection{Qualitative results}
Figure~\ref{fig:compare_results} shows generated images using our method, as well as baselines. From these examples it is clear that our method can generate complex images with multiple objects, and even multiple instances of the same object type. For example, Figure~\ref{fig:compare_results}(a) shows two boats, (c) shows two cows, (e) and (r) contain two people.

These examples also show that our method generates images which respect the location constraints of the input bounding boxes, and the generated objects in the image are also recognizable and consistent with their input labels.

As we can see in Figure~\ref{fig:compare_results}, pix2pix fails to generate meaningful images, due to the extreme difficulty of directly mapping layout to a real image without detailed instance segmentation. The results generated by sg2im are also not as good as ours. For example, in Figure~\ref{fig:compare_results} (g) and (i), the generated giraffe and zebra are difficult to recognize, and (l) contains lots of artifacts, making result look unrealistic.

In Figure~\ref{fig:adding_results} we demonstrate our model's ability to generate complex images by starting with simple layout and progressively adding new bounding box or moving existing bounding box, \eg, (g) and (k), to build/manipulate a complex image. From these examples we can see that new objects are drawn in the images at the desired locations, and existing objects are kept consistent as new content is added.

Figure~\ref{fig:diverse_results} shows the diverse results generated from the same layouts. Given that the same layout may have many different possible real image realizations, the ability to sample diverse images is a key advantage of our model. 


\subsection{Quantitative results}



Table~\ref{tb:quantitative_results} summarizes comparison results of the inception score, object classification accuracy and diversity score of baseline models and our model. 
We also report the inception score and object classification accuracy on real images.

The proposed method significantly outperforms baselines in all the three evaluation metrics.
In terms of Inception Score and Fr\'echet Inception Distance, our method outperforms the existing approaches with a substantial margin, presumably because our method generates more recognizable objects as proved by object classification accuracy.
Please note that the object accuracy on real images is not the upper bound of object classification accuracy, since the object cannot be classified correctly in a real image does not necessarily mean it is also difficult to distinguish in a generated image.
Since the pix2pix is deterministic, its diversity score is 0.
By adding global noise to scene layout, sg2im can generate images with limited diversity.
The diversity performance shows that our method can generate diverse results from the same layout. A very notable improvement is on COCO, where we achieve diversity score of 0.15 as compared to 0.02 for sg2im.

\begin{table}[!t]
\centering
\begin{tabular}{l|ccc}
\textbf{Method}     & \textbf{IS}   & \textbf{Accu.}  & \textbf{DS}   \\\hline 
w/o $\mathcal{L}_1^\mathrm{img }$      & 7.6 $\pm$ 0.2   & 49.03 & 0.17 $\pm$ 0.09 \\
w/o $\mathcal{L}_1^\mathrm{latent}$   & 7.5 $\pm$ 0.1   & 48.90 & 0.16 $\pm$ 0.09 \\
w/o $\mathcal{L}_\mathrm{AC}^\mathrm{obj}$        & 6.5 $\pm$ 0.1   & 10.06 & \textbf{0.37 $\pm$ 0.11} \\
w/o $\mathcal{L}_\mathrm{adv}^\mathrm{img}$ & 7.1 $\pm$ 0.1 & 56.17 & 0.13 $\pm$ 0.09\\
w/o $\mathcal{L}_\mathrm{adv}^\mathrm{obj}$ & 7.3 $\pm$ 0.1 & \textbf{57.74} & 0.14 $\pm$ 0.09 \\\hline
full model         & \textbf{8.1} $\pm$ \textbf{0.1} & 48.09 &  0.17 $\pm$ 0.09 \\ 
\end{tabular}
\vspace{0.1in}
\caption{\textbf{Ablation study of our model} on Visual Genome dataset by removing different objectives. IS is the inception score, Accu. is the object classification accuracy, and DS is the diversity score.}
\label{tb:ablation}
\vspace{-0.2in}
\end{table}


\subsection{Ablation Study}
We demonstrate the necessity of all components of our model by comparing the inception score, object classification accuracy, and diversity score of several ablated versions of our model trained on Visual Genome dataset:
\begin{itemize}[leftmargin=*]
\setlength{\itemsep}{0pt}
	\item \textbf{w/o $\mathcal{L}_1^\mathrm{img}$} reconstructs ground truth images without pixel regression.
	\item \textbf{w/o $\mathcal{L}_1^\mathrm{latent}$} does not regress the latent codes which are used to generated objects in the result images.
	\item \textbf{w/o $\mathcal{L}_\mathrm{AC}^{\mathrm{obj}}$} does not classify the category of objects.
	\item \textbf{w/o $\mathcal{L}_\mathrm{adv}^\mathrm{img}$} removes the object adversarial loss when training the model.
	\item \textbf{w/o $\mathcal{L}_\mathrm{adv}^\mathrm{obj}$} removes the image adversarial loss when training the model.
\end{itemize}

As shown in Table~\ref{tb:ablation}, removing any loss term will decrease the overall performance.
Specifically, The model trained without $\mathcal{L}_1^\text{img}$ or $\mathcal{L}_1^\text{latent}$ generates less realistic images, which decreases the inception score. The object classification accuracy is still high because of the object classification loss.
Without the constraint on reconstructed images or latent codes, the models get lower inception scores, but similar diversity scores.
Removing the object classification loss degrade the inception score and object classification accuracy significantly, since the model cannot generate recognizable objects. Not surprisingly, this freedom results in higher diversity score.
It is expected to see that removing the adversarial loss on image or object will decrease the inception score substantially. 
However, the object classification accuracy increases further comparing to the full model. We believe that without the realism requirement of image or object, the object classification loss could be tampered with adversarial attack.  
Trained with all the losses, our full model achieves a good balance across all three metrics.

\section{Conclusion}
In this paper we have introduced an end-to-end method for generating diverse images from layout (bounding boxes~+ categories).
Our method can generate reasonable images which look realistic and contain recognizable objects at the desired locations.
We also showed that we can control the image generation process by adding/moving objects in the layout easily.
Qualitative and quantitative results on COCO-Stuff~\cite{caesar2016coco} and Visual Genome~\cite{krishna2017visual} datasets demonstrated our model's ability to generate realistic complex images.
Generating high resolution images from layouts will be our future work. 
Moreover, making the image generation process more controllable, such as specifying the fine-grained attributes of instances, would be an interesting future direction.

\vspace{+0.1in}
\small \noindent \textbf{Acknowledgement}
This research was supported, in part, by NSERC Discovery, NSERC DAS and NSERC CFI grants. 
We gratefully acknowledge the support of NVIDIA Corporation with the donation of the Titan V GPU used for this research.
{\small
\bibliographystyle{ieee_fullname}
\bibliography{bib}
}

\end{document}